\pdfoutput=1
\documentclass[11pt]{article}
\usepackage{multirow}
\usepackage[]{koti2023}

\usepackage{times}
\usepackage{latexsym}

\usepackage[T1]{fontenc}
\usepackage[utf8]{inputenc}

\usepackage{microtype}

\usepackage{inconsolata}

\usepackage{graphicx}
\graphicspath{ {.images/} }

\title{Keyword-Optimized Template Insertion for Clinical Note Classification via Prompt-Based Learning}

\author{Eugenia Alleva \\
  Windreich Department of Artificial Intelligence  and Human Health at Mount Sinai \\
  Hasso Plattner Institute for Digital Health at Mount Sinai \\Icahn School of Medicine at Mount Sinai, New York, USA \\
  \texttt{eugeniaalessandrae.allevabonomi@mssm.edu} 
  \AND
  Isotta Landi\\
   Institute for Personalized Medicine \\ Icahn School of Medicine at Mount Sinai, New York, USA \\
   \AND
  Leslee J Shaw\\
   Blavatnik Family Women's Health Research Institute \\ Icahn School of Medicine at Mount Sinai, New York, USA \\
   \AND
   Thomas J Fuchs \\
  Windreich Department of Artificial Intelligence  and Human Health at Mount Sinai \\
  Hasso Plattner Institute for Digital Health at Mount Sinai \\Icahn School of Medicine at Mount Sinai, New York, USA \\
  \AND
  Erwin Böttinger\\
   Hasso Plattner Institute for Digital Health at Mount Sinai \\ Icahn School of Medicine at Mount Sinai, New York, USA \\
  \AND
  Ipek Ensari \\
   Windreich Department of Artificial Intelligence  and Human Health at Mount Sinai \\
  Hasso Plattner Institute for Digital Health at Mount Sinai \\ 
  Blavatnik Family Women's Health Research Institute at Mount Sinai
  \\Icahn School of Medicine at Mount Sinai, New York, USA \\}

\begin{document}
\maketitle
\begin{abstract}

Clinical note classification is a common clinical NLP task. However, annotated data-sets are scarse. Prompt-based learning has recently emerged as an effective method to adapt pre-trained models for text classification  using only few training examples. A critical component of prompt design is the definition of the template (i.e. prompt text). The effect of template position, however, has been insufficiently investigated. This seems particularly important in the clinical setting, where task-relevant information is usually sparse in clinical notes. In this study we develop a keyword-optimized template insertion method (KOTI) and show how optimizing position can improve performance on several clinical tasks in a zero-shot and few-shot training setting.   
\end{abstract}

\section{Introduction}

Clinical note classification is a common clinical NLP task, and often a necessary step to correctly characterize patient cohorts from electronic health records. 
The use of large models, pre-trained on clinical text, and adapted to the specific task via standard fine-tuning is, however, limited by the small size of annotated datasets available. This limitation is not surprising considering the fact that annotation requires extensive domain knowledge which, in the clinical field, is particularly expensive \citep{annotation_expensive}. 

Prompt-based learning \citep{prompt_survey, neurips, manual_verb, PBL_start} has recently emerged as an effective technique to adapt pre-trained models with no or only few  training examples. In its simplest form, a prompting text (i.e. template), which includes a [MASK] token, is appended to a model's input. This conditions the model's output for the [MASK] token such that it can be directly mapped to a label class through a verbalizer  \citep{manual_verb}. While several aspects of prompt design have been extensively investigated, only few studies have characterized the effect of template position. This is a particularly interesting question in the context of clinical note classification, where relevant information is usually sparse.

In this work we assessed the effect of template position in a zero-shot and few-shot prompt-based learning setting for clinical note classification. This takes the form of three contributions
\begin{itemize}
    \item We developed a keyword-optimized template insertion method (KOTI), to identify optimal template positions and show that KOTI can improve model performance in zero-shot and few-shot training setting.
    \item We demonstrate that, with standard template insertion, truncating the clinical note so as to favor the presence of keywords can improve performance over naive 'tail' truncation, while maintaining computational efficiency. 
    \item We show that in few-shot learning, training on balanced examples improves performance over training on a random samples of the same size.
\end{itemize}

\section{Related Work}
 
\paragraph{Prompt-Based Learning for Clinical Note Classification}
Several authors have explored the use of prompt-based learning with encoder models for clinical note classification.  
 \citet{healthprompt} build a framework for classifying clinical notes via prompt-based learning in a zero-shot setting, achieving high performance for ICD-10 disease codes classification tasks. 
 \citet{clinicalPromptFrozen} compare several combinations of manual and soft (i.e. learnable) templates and verbalizers to classify notes for ICD-10 codes. During training, they update prompt-related parameters while keeping the pre-trained language model frozen, and report a better performance with respect to standard fine-tuning.  \citet{dementia_prompt} train a prompt-based classifier for dementia detection.  \citet{longformerICDknowledge} apporach clinical note classification for ICD-9 codes as a multilabel task, simultaneously appending one prompt template per ICD-9 code. They also include a step of knowledge injection via contrastive learning using usmle and ICD9 hierarchical codes. 

The effect of template position is only investigated by   \citet{dementia_prompt}, where they compare appending and pre-pending the template without showing a consistent advantage of one over the other.  \citet{healthprompt} mention the problem of clinical note trucation due to token number limitations. In their framework, they solve this issue by splitting notes into smaller chunks, labels estimated for each chunk and finally aggregated via max pooling. 

\paragraph{Prompt Template Position}
Only few studies have investigated the effect of template position, all of them concentrating on soft templates only.  \citet{soft_prompt_pos} investigate the effect of soft template position (appending vs pre-pending) for single-sentence or sentence-pair classification without splitting the original input sentence, while  \citet{prompt_position} also interspersed the template tokens across the sentence. Both show how prompt template position impacts model performance, while suggesting that optimal template position is task-dependent.  Finally,  \citet{dynamic_prompt} develop a dynamic prompting framework, where soft template position is optimized together with length and representation, considering only appending and pre-pending as insertion methods. 

\section{Methods}
\subsection{Prompt design}
To design our prompt templates and verbalizers we adapted the OpenPrompt framework  \citep{openprompt}.
We simulated a situation with minimal prompt engineering, directly translating the task definition and labels into prompt templates and verbalizers. For all tasks we used the following design principles. A complete list of templates can be found in Table \ref{prompt_table}.

\textbf{Prompt template}
We used prefix-type manual templates of the form \texttt{<task objective> : [MASK]}

\textbf{Verbalizer}
We used a manual verbalizer as first described by  \citet{manual_verb}. In brief, a set of label words is selected for each class. We then perform normalization and softmax over the label word logits for the model's prediction at the [MASK] token to identify the predicted class.  We selected one single label word per label class. (See Table \ref{prompt_table}).  

\begin{table*}[h]
\centering
\resizebox{\textwidth}{!}{
\begin{tabular}{p{0.08\textwidth} p{0.25\textwidth} p{0.3\textwidth} p{0.23\textwidth} p{0.35\textwidth}}
\hline
    \textbf{Task} & \textbf{Classes} &\textbf{Template} & \textbf{Label-Words} & \textbf{Keywords} \\
\hline
    \textbf{Dys} & Yes, No, Unknown & dysmenorrhea: [MASK] & yes, no, unknown & dysmenorrhea, cramps, menstrual pain, period pain  \\
    \textbf{OA} & Yes, Unmentioned & osteoarthritis (OA): [MASK] & yes, no & depressive, depression, mood  \\
    \textbf{Dep} & Yes, Unmentioned &  depression: [MASK] & yes, no & bone, osteo, arthritis, osteoarthritis, joint, cartilage, OA  \\
    \textbf{PVD} & Yes, Unmentioned & peripheral vascular disease (PVD): [MASK] & Yes, No & vascular, peripheral vascular, arterial \\
    \textbf{Smk} & current, past, no, unknown & smoking: [MASK] & yes, no, past, unknown & smoking, smoke, cigar, cigarette \\
\hline
\end{tabular}
}
    \caption{Prompt configuration for each task.}
    \label{prompt_table}
\end{table*}

\subsection{Template position}
\subsubsection{Keyword-Optimized Template Insertion}
In KOTI, salient regions of the clinical note are identified through keyword-matching, and the prompt template inserted in their vicinity (Appendix \ref{appendix prompt design}). For each tasks, we performed the following steps: 
\begin{enumerate}
\item \textbf{Keyword Definition} We first built a set of task-specific keywords which are semantically and clinically closely related to the task specification (see Table \ref{prompt_table}).
\item \textbf{Template Insertion}: We then identified sentences within clinical notes and flagged them as containing one or more keywords. We split the clinical note \texttt{<text>} at the end of the first flagged sentence into two sub-chunks, \texttt{<text\_a>} and \texttt{<text\_b>}, and insert the prompt template between them. (\texttt{<text\_a> <template> <text\_b>})
\item \textbf{Clinical Note Truncation} Clinical notes usually contain more tokens than what accepted by the models. We therefore performed head truncation on \texttt{<trim\_text\_a>} and tail truncation on \texttt{<text\_b\_trim>}, with the number of truncated tokens proportional to the sub-chunk's lengths. The final input is therefore \texttt{<trim\_text\_a> <template> <text\_b\_trim>}
\end{enumerate}

\subsubsection{Standard Template Insertion}
We compare KOTI with standard template insertion (STI) at the end of the input text. The truncation method employed in KOTI indirectly selects a "salient" text chunk as the model's input. We therefore employ and compare two different standard template insertion situations:

\textbf{Standard Chunk (STI-s)} The clinical note  \texttt{<text>} taken as is and tail truncated to fit into the model input (\texttt{<text\_trim>}). 

\textbf{Keyword Chunk (STI-k)}The same sub-chunks used in KOTI, i.e.  \texttt{<trim\_text\_a>} and \texttt{<text\_b\_trim>}, are concatenated and the template text appended at the end of the input.

\subsection{Pretrained Model and Prompt-based Fine-tuning}
We compared the performance of two encoder models, GatorTron  \citep{gatortron} and ClinicalBERT  \citep{clinicalbert}. Both models where trained at least partially on clinical notes. We performed prompt-based model fine-tuning as described in  \citet{manual_verb}. In brief, we fine-tuned all model parameters by minimizing the cross entropy loss between the verbalizer's probability output and the true label.  

\subsection{Tasks}
We performed experiments on five different classifications tasks (Table \ref{prompt_table}). A detailed description of each task can be found in Appendix \ref{appendix_tasks}. 
In Brief, we used two publicly available data-sets, the N2C2 smoking challenge data-set (\textbf{Smk}) for smoking status classification task, and the N2C2 obesity challenge data-set for  peripheral vascular disease (\textbf{PVD}), depression (\textbf{Dep}) and osteoarthritis (\textbf{OA} classification tasks. For both challenges, clinical notes consisted of discharge summaries. Moreover, we included our own task for classification of dysmenorrhea (\textbf{Dys}) from clinical notes of gynecological encounters. Note selection and manual annotation was performed as described in Appendix \ref{appendix_tasks}. 

\subsection{Experimental Setup}
For each task, we compared the performance of KOTI, STI-k and STI-s in different training setting:

\textbf{Zero-Shot} We evaluated out-of-the-box performance of the models on the validation data-set.

\textbf{Few-Shot with Balanced Examples} We evaluated the performance of the models trained on k=1, 4, and 10 examples per label class.

\textbf{Few-Shot with Random Examples} We evaluated the performance of the models trained on as many examples as in the balanced setting, but randomly sampled (i.e. reflecting the natural label class distrubtion). Moreover, we included runs on 50 and 100 training examples.

\paragraph{Hyper-parameter optimization}
We optimized model hyper-parameters for each combination of task, training configuration, model and template insertion method via random search (See appendix \ref{appendix_training} for more details). 
We performed 10 runs for each hyper-parameter combination, where we randomly selected training examples from the train data-set, evaluating performance on the remaining sample. Average F1/Macro-F1 scores where used to select optimal hyper-parameters.

\paragraph{Testing}
For each training setting, we simulated a real-world scenario and randomly sampled k number of training examples from the validation data-set, evaluating the model's performance on the remaining examples. We repeated this procedure for 10 runs to estimate average precision, recall and F1 scores. We report MacroF1 scores for multi-class tasks.

\section{Results}
The complete results can be found in Appendix \ref{appendix_results}.
Table \ref{table_best_res} shows the best models for each task and their performances. KOTI consistently leads to best model with exception of PVD task with GatoTron. 

\begin{table} [h]
\centering 
\resizebox{0.45\textwidth}{!}{
\begin{tabular}{p{0.05\textwidth} p{0.11\textwidth} p{0.10\textwidth} p{0.10\textwidth} p{0.05\textwidth} p{0.10\textwidth}}
\hline
\textbf{Task}                 & \textbf{Model} & \textbf{Template} & \textbf{Average F1/Macro-F1} & \textbf{N} & \textbf{Sampling} \\
\hline
\multirow{2}{*}{\textbf{Dep}} & GatorTron      & KOTI              & 0.919                        & 100        & random            \\
                              & ClinBERT       & KOTI              & 0.851                        & 100        & random            \\
\multirow{2}{*}{\textbf{OA}}  & GatorTron      & KOTI              & 0.702                        & 100        & random            \\
                              & ClinBERT       & KOTI              & 0.667                        & 4          & balanced          \\
\multirow{2}{*}{\textbf{PVD}} & GatorTron      & STI-k             & 0.767                        & 100        & random            \\
                              & ClinBERT       & KOTI              & 0.592                        & 100        & random            \\
\multirow{2}{*}{\textbf{Dys}} & GatorTron      & KOTI              & 0.913                        & 100        & random            \\
                              & ClinBERT       & KOTI              & 0.820                        & 100        & random            \\
\multirow{2}{*}{\textbf{Smk}} & GatorTron      & KOTI              & 0.620                        & 100        & random            \\
                              & ClinBERT       & STI-k             & 0.510                        & 100        & random     
\\
\hline
\end{tabular}
}
    \caption{Best Model for Each Task.}
    \label{table_best_res}
\end{table}

\paragraph{Zero-shot} Table \ref{table_0S} shows the performance of the models in zero-shot setting for every configuration. KOTI outperforms STI-k and STI-s in every task with both ClinicalBERT and GatorTron.

\begin{table}[h]
\centering
\resizebox{0.30\textwidth}{!}{
\begin{tabular}{p{0.05\textwidth} p{0.10\textwidth} p{0.07\textwidth} p{0.07\textwidth} p{0.07\textwidth}}
\hline
\textbf{Task}                 & \textbf{Model} & \textbf{KOTI} & \textbf{STI-k} & \textbf{STI-s} \\
\hline
\multirow{2}{*}{\textbf{Dep}} & GatorTron      & 0.358         & 0.034          & 0.065          \\
                              & ClinBERT       & 0.012         & 0.000          & 0.000          \\
\multirow{2}{*}{\textbf{OA}}  & GatorTron      & 0.277         & 0.020          & 0.044          \\
                              & ClinBERT       & 0.081         & 0.000          & 0.000          \\
\multirow{2}{*}{\textbf{PVD}} & GatorTron      & 0.386         & 0.000          & 0.030          \\
                              & ClinBERT       & 0.000         & 0.000          & 0.000          \\
\multirow{2}{*}{\textbf{Dys}} & GatorTron      & 0.414         & 0.240          & 0.156          \\
                              & ClinBERT       & 0.206         & 0.171          & 0.154          \\
\multirow{2}{*}{\textbf{Smk}} & GatorTron      & 0.141         & 0.117          & 0.127          \\
                              & ClinBERT       & 0.250         & 0.100          & 0.076  
                              \\
\hline
\end{tabular}
}
\caption{Zero-Shot Performance. We report average F1/Macro-F1 scores. See Appendix \ref{appendix_results} for more results.}
\label{table_0S}
\end{table}

\paragraph{Few-shot} Figure \ref{fig1} shows the average performance across tasks for each experimental setting. On average, KOTI improves performance in most few-shot scenarios with respect to STI-k, while STI-s shows the worst performance. Moreover, training with balanced examples improves model performance with respect to training on random examples.

\section{Discussion}
\paragraph{KOTI} The impact of template position has mostly been neglected in prompt-based learning research. In this work we demonstrate that, for zero-shot clinical note classification, KOTI improves model performance over both STI-k and STI-s. Because the same input text chunk is used in both STI-k and KOTI, this effect is driven by template position alone. Moreover, we find that KOTI can lead to significant improvements over STI-k in a few-shot scenario, especially when training sampels are unbalanced. This advantage, however, is more variable across different tasks (See Appendix \ref{appendix_results}), corroborating the hypothesis that effect of template position is task-dependent, as reported by  \citet{prompt_position}. This variability might also reflect different performances of selected templates, verablizers and keywords. We did not engineer them through trial-and-error but rather selected them \textit{a priori} (with the  exception of Dys keywords). A different approach might be more suitable.
\begin{figure}[h]
\includegraphics[width=0.5\textwidth]{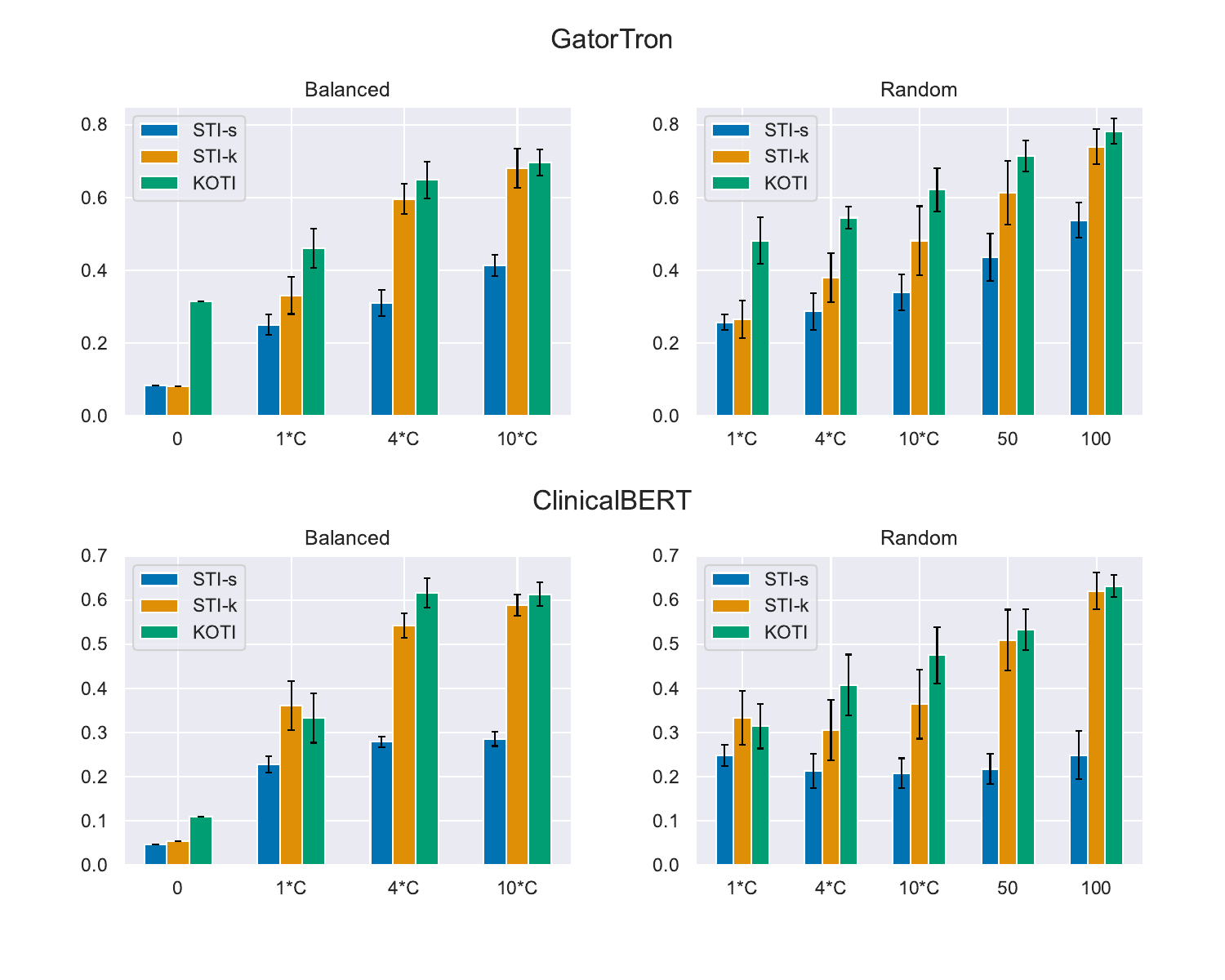}
\caption{Average Performance Across Tasks. F1 and Macro-F1 scores across the tasks are averaged and reported with standard errors. C=number of classes}. 
\label{fig1}
\end{figure}

\paragraph{Keyword Chunks} Due to input size constraints, we often need to truncate the clinical note. Because relevant information within the clinical notes is sparse, performing this process 'naively' can lead to loss of the most valuable text. It is therefore no surprise that, on average, STI-s performs worse than both KOTI and STI-k in all settings. Loss of valuable information seems to be particularly detrimental with ClinicalBERT, where we can appreciate a stagnation of performance as a function of training examples (Fig.\ref{fig1}). The HealthPrompt framework  \citep{healthprompt} avoids this issue by chunking the text into smaller pieces, processing each chunk separately. Consistently with our results, they show that without chunking, performance drops dramatically  \citep{healthPromptEval}. Their approach, however, requires multiple inference runs per clinical note. In the N2C2 obesity data-set, for example, the 98\% of the notes are longer than the 512 token limit of both our models, and would require on average 3 times more inference run with separate chunk processing (See Appendix \ref{appendix_results}). Our keyword-based approach, instead, is computationally more efficient, requiring only 1 run per note.

\paragraph{KOTI and Unbalanced Training Examples} Finally, our results provide evidence that when label classes are unbalanced, training with balanced examples in a few-shot setting can improve model performance. In this regards, KOTI seems to provide a consistent advantage when only few random examples are available for training. Indeed, Fig. \ref{fig1} suggests that, with KOTI, the models might learn faster, reaching a higher performance with fewer examples, while converging on performances similar to STI-k when more examples are available. This could reflect KOTI's higher zero-shot performance, which might improve prediction of rare classes with few training examples.

\section*{Limitations}
We report several limitations of our work. Firstly, we developed KOTI to work on encoder architectures and assessed its effect for this specific scenario. Our results do not inform on the utility of KOTI or related methods in the context of decoders. Our decision to focus on encoders mainly relies on the fact that most pre-trained models adapted to clinical notes are encoders  \citep{clicalLMsurvey}, with few exceptions  \citep{clinicalT5}. Secondly, our method requires some engineering to select effective keywords. While we show promising results using \textit{a priori} keywords selection with limited engineering, we achieved the best results on the dysmenorrhea task for which keywords were more accurately selected. Moreover, keyword matching requires some degree of computational efforts for note processing and text pattern matching. Further studies should assess the benefits KOTI over full note processing such as in  \citet{healthprompt}. Lastly, we assess KOTI only in the clinical domain. It remains unclear wether the same gains in performance would be seen in other NLP domains.

\section*{Ethics Statement}
This work uses deidentified documents from the publicly available n2c2 NLP Research Data Sets. The use of the notes for the dysmenorrhea task dataset was approved by the hospital system IRB.

\bibliography{koti2023}
\bibliographystyle{acl_natbib}

\appendix

\section{Prompt Design}
\begin{figure}[h]
\includegraphics[width=0.5\textwidth]{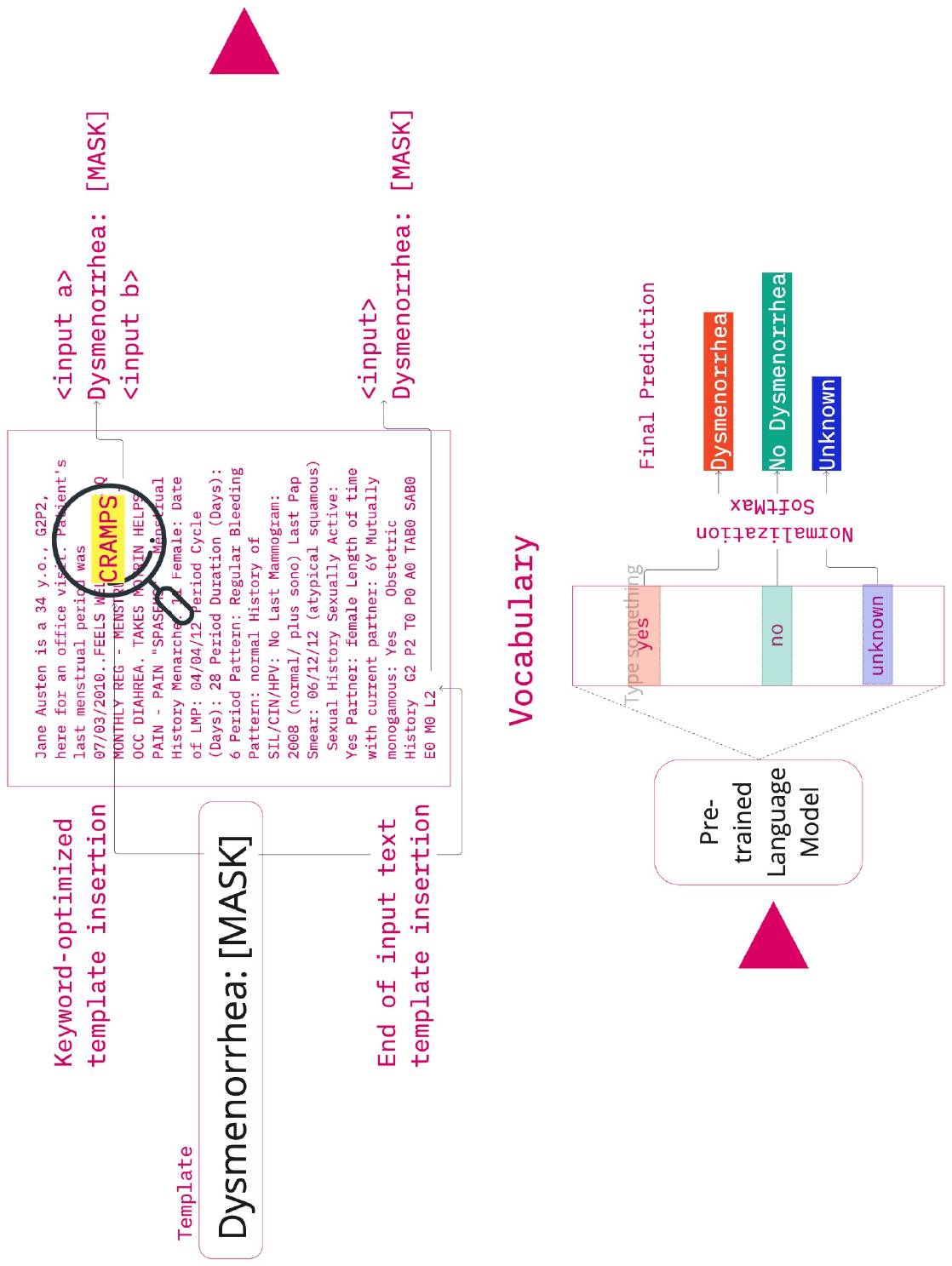}
\caption{Keyword-Optimized Template Insertion}. 
\label{fig_pr}
\end{figure}

Figure \ref{fig_pr} schematizes our prompt design. 

Using the keywords reported in Table \ref{prompt_table}, we identified salient sentences in 0.21 notes for osteoarthritis, 0.24 notes for peripheral vascular disease, 0.18 notes for depression, 0.20 notes for smoking, and 0.58 notes for dysmenorrhea.

\label{appendix prompt design}

\section{Tasks}
\label{appendix_tasks}

\paragraph{Dysmenorrhea Classification (Dys)} The aim of this task was to classify clinical notes for \textit{dysmenorrhea}, \textit{no dysmenorrhea} or \textit{unknown}. For this task, we randomly selected and manually annotated 300 clinical notes related to routine gynecological examinations from a large multi-center hospital system data warehouse, and equally split them into train and validation set. Annotation was performed by a medical doctor with experience in gynecology. 

\paragraph{N2C2 Obesity Challenge - Co-morbidities}
The N2C2 Obesity Challenge dataset  \citep{obesity_challenge} consists of 1237 discharge summaries of obese or overweight patients and annotated for obesity and a list of co-morbidities. Annotations are either textual, for information that is explicitly written within notes, or intuitive, reflecting domain expert medical professionals' reading of the information presented. We selected a subset of co-morbidities with binary label classes and representing diverse clinical specialties: \textbf{Osteoarthitis (OA)}, \textbf{Depression (Dep)} and \textbf{Peripheral Vascular Disease (PVD)}. We performed experiments for each of these sub-tasks separately, using annotations for textual information as ground truth. 

\paragraph{N2C2 Smoking Challenge (Smk)}
The N2C2 Smoking Challenge  \citep{smoking_challenge} aims at identifying smoking status from clinical notes. The dataset consists of 502 discharge summaries annotated for smoking status with 5 classes: smoker, current smoker, past smoker, no smoker, unknown. Because the smoker class had only 3 examples in the test dataset, we merged current and smoker into a single current smoker class. 

Table \ref{task_table} shows the distribution of the labels classes in each task for training and test sets.

\begin{table}[]
\begin{tabular}{llll}
\hline
\textbf{Task}                 & \textbf{Class} & \textbf{Train} & \textbf{Validation} \\
\hline
\multirow{3}{*}{\textbf{Dys}} & Yes            & 34             & 37                  \\
                              & No             & 52             & 45                  \\
                              & Unknown        & 64             & 68                  \\
\multirow{2}{*}{\textbf{Dep}} & Yes            & 86             & 72                  \\
                              & Unknown        & 519            & 434                 \\
\multirow{2}{*}{\textbf{PVD}} & Yes            & 83             & 65                  \\
                              & Unknown        & 526            & 443                 \\
\multirow{2}{*}{\textbf{OA}}  & Yes            & 89             & 86                  \\
                              & Unknown        & 514            & 416                 \\
\multirow{4}{*}{\textbf{Smk}} & Current Smoker & 44             & 14                  \\
                              & Past Smoker    & 36             & 11                  \\
                              & No Smoker      & 66             & 16                  \\
                              & Unknown        & 252            & 63  
                              \\
\hline
\end{tabular}
\caption{Class distribution for different tasks.}
\label{task_table}
\end{table}

\section{Hyper-parameter tuning and training}
\label{appendix_training}

Hyper-parameter where optimized via 10 trials of random search over the following hyper-parameter space :
\begin{itemize}
    \item Learning Rate over a random uniform distribution on [1e-4; 1e-7]
    \item Batch size over the set {1,2,4}
    \item Number of epochs over the set {1,2,3,4,5,6,7,8,9,10}
\end{itemize}
The optimal hyper-parameters where selected based of F1 scores for binary tasks and MacroF1 scores for multiclass tasks. 

Training of the models was performed on NVIDIA a100 GPUs.

\section{Complete Results}
\label{appendix_results}

\begin{figure}[h]
\includegraphics[width=0.5\textwidth]{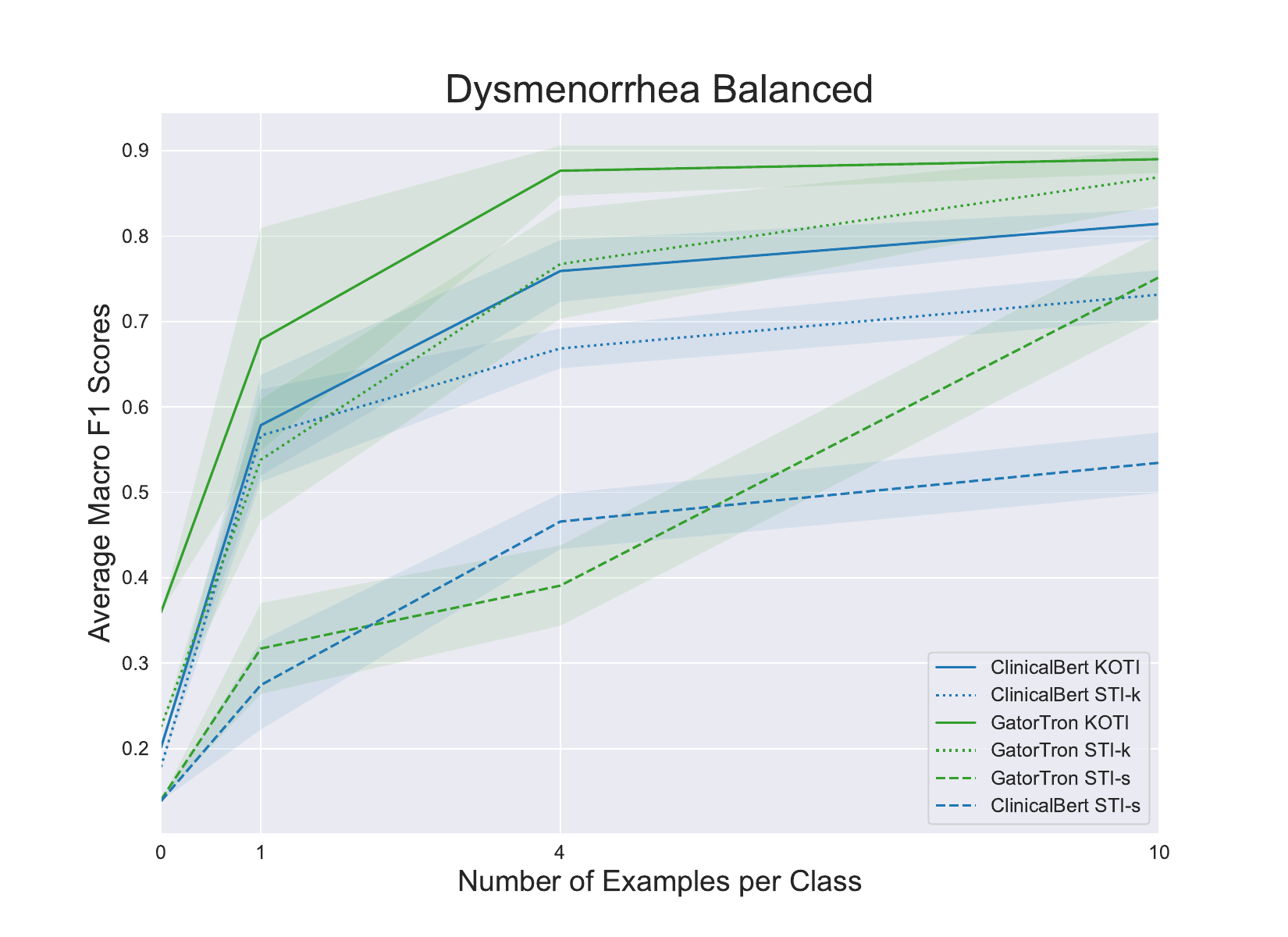}
\caption{Dysmenorrhea task with balanced examples}
\label{dys_b}
\end{figure}

\begin{figure}[h]
\includegraphics[width=0.5\textwidth]{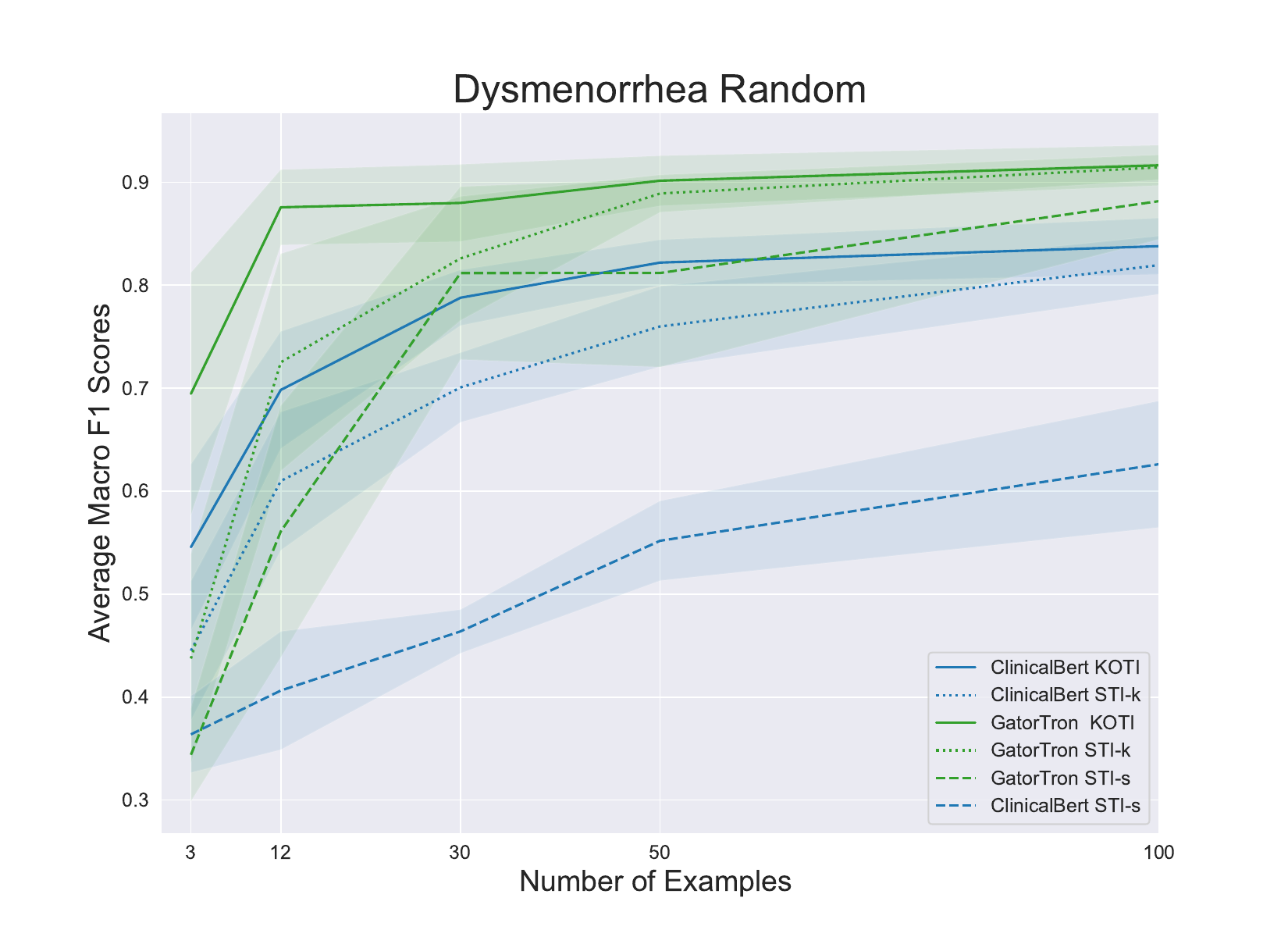}
\caption{Dysmenorrhea task with random examples}
\label{dys_r}
\end{figure}

\begin{figure}[h]
\includegraphics[width=0.5\textwidth]{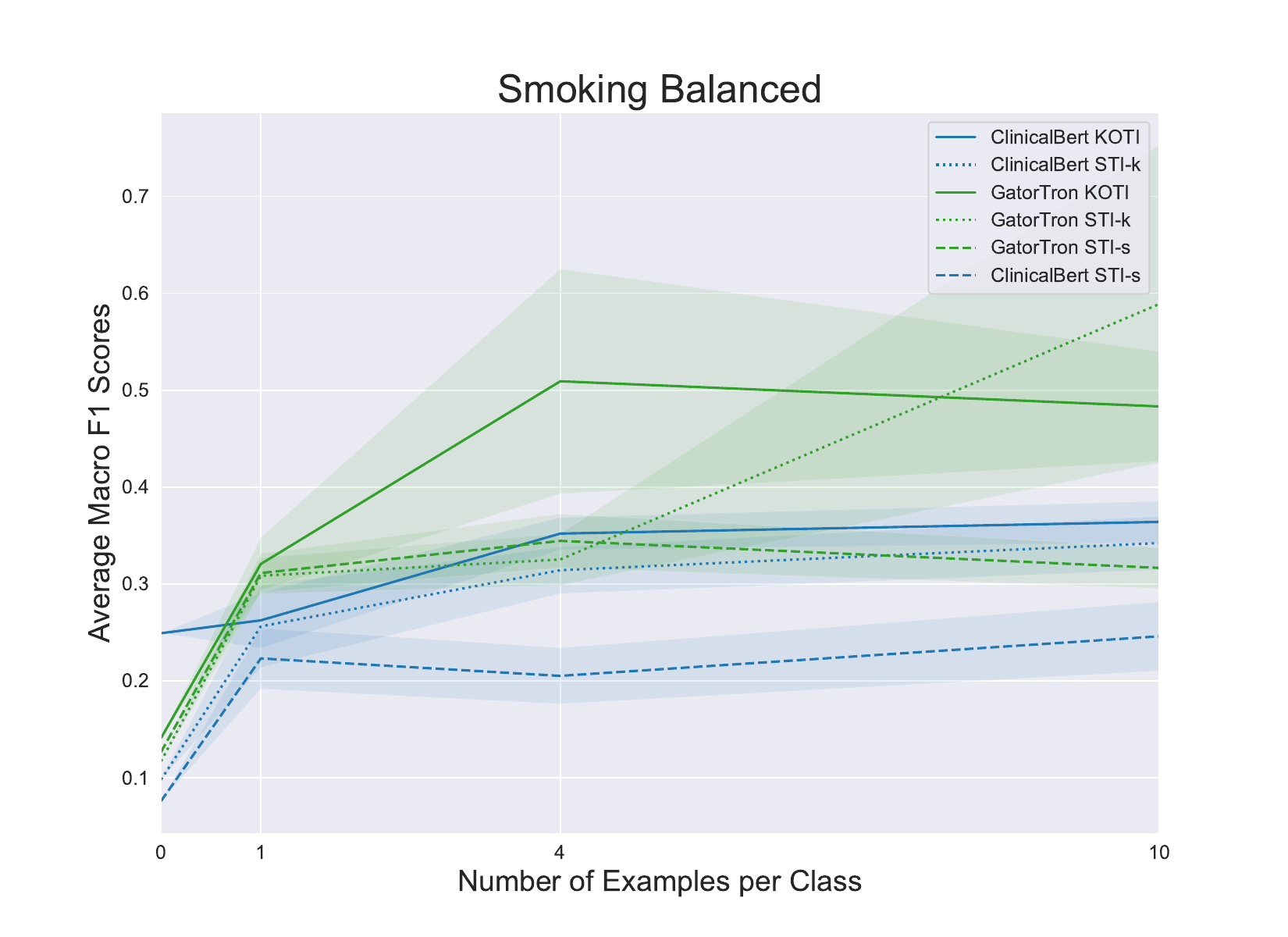}
\caption{Smoking task with balanced examples}
\label{smk_b}
\end{figure}

\begin{figure}[h]
\includegraphics[width=0.5\textwidth]{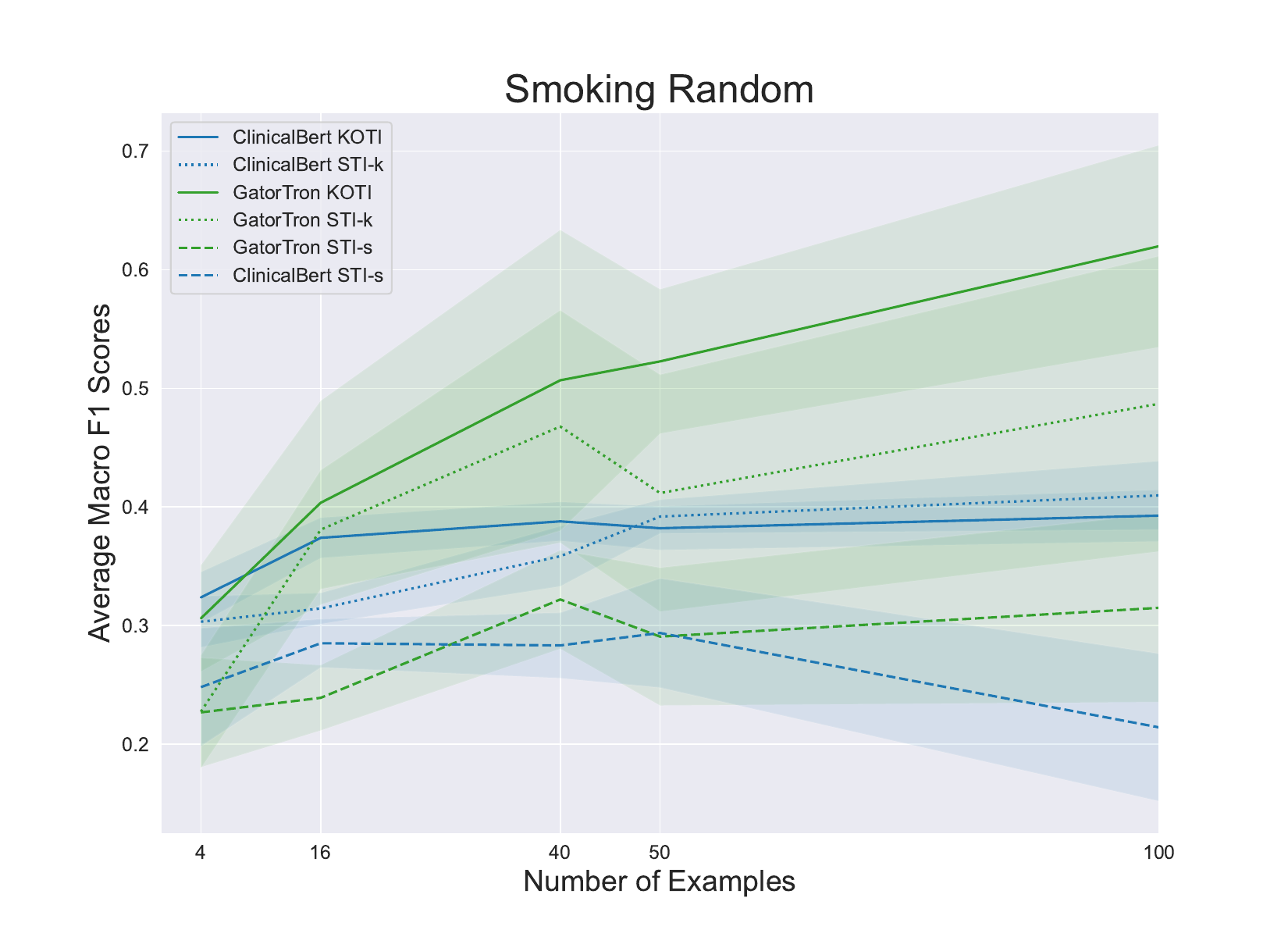}
\caption{Smoking task with random examples}
\label{smk_r}
\end{figure}

\begin{figure}[h]
\includegraphics[width=0.5\textwidth]{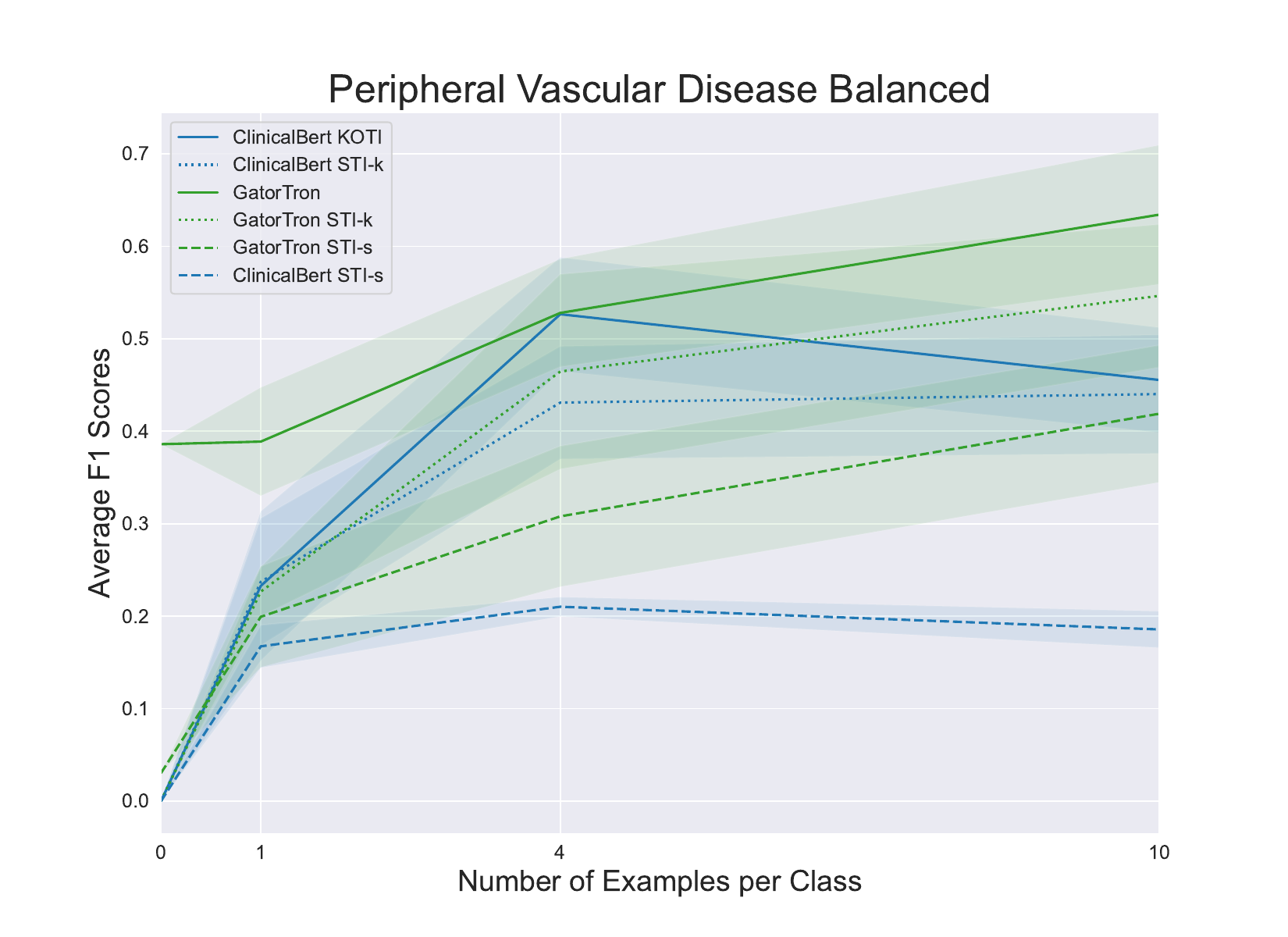}
\caption{Peripheral vascular disease task with balanced examples}
\label{pvd_b}
\end{figure}

\begin{figure}[h]
\includegraphics[width=0.5\textwidth]{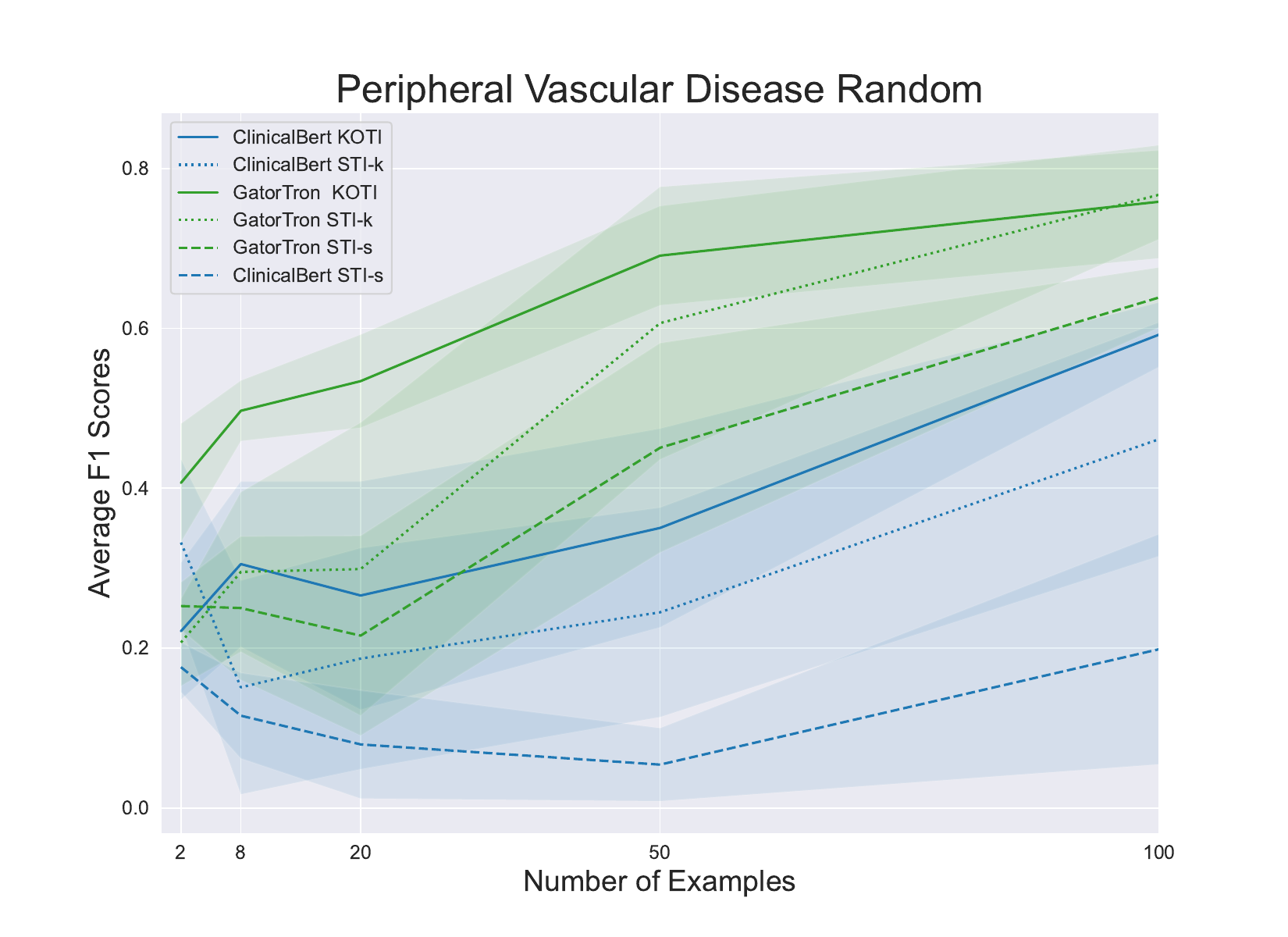}
\caption{Peripheral vascular disease  task with random examples}
\label{pvd_e}
\end{figure}

\begin{figure}[h]
\includegraphics[width=0.5\textwidth]{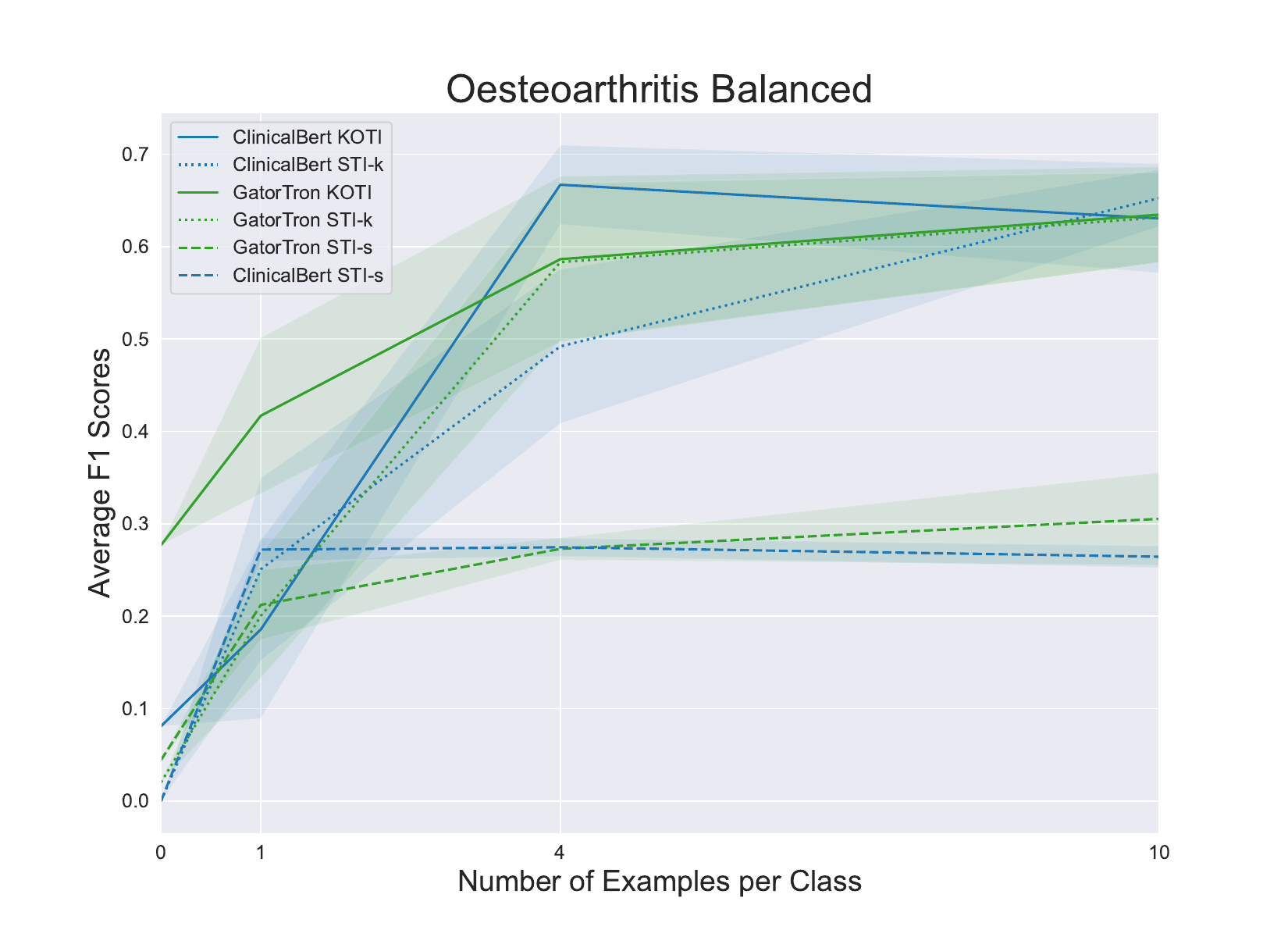}
\caption{Depression task with balanced examples}
\label{oa_b}
\end{figure}

\begin{figure}[h]
\includegraphics[width=0.5\textwidth]{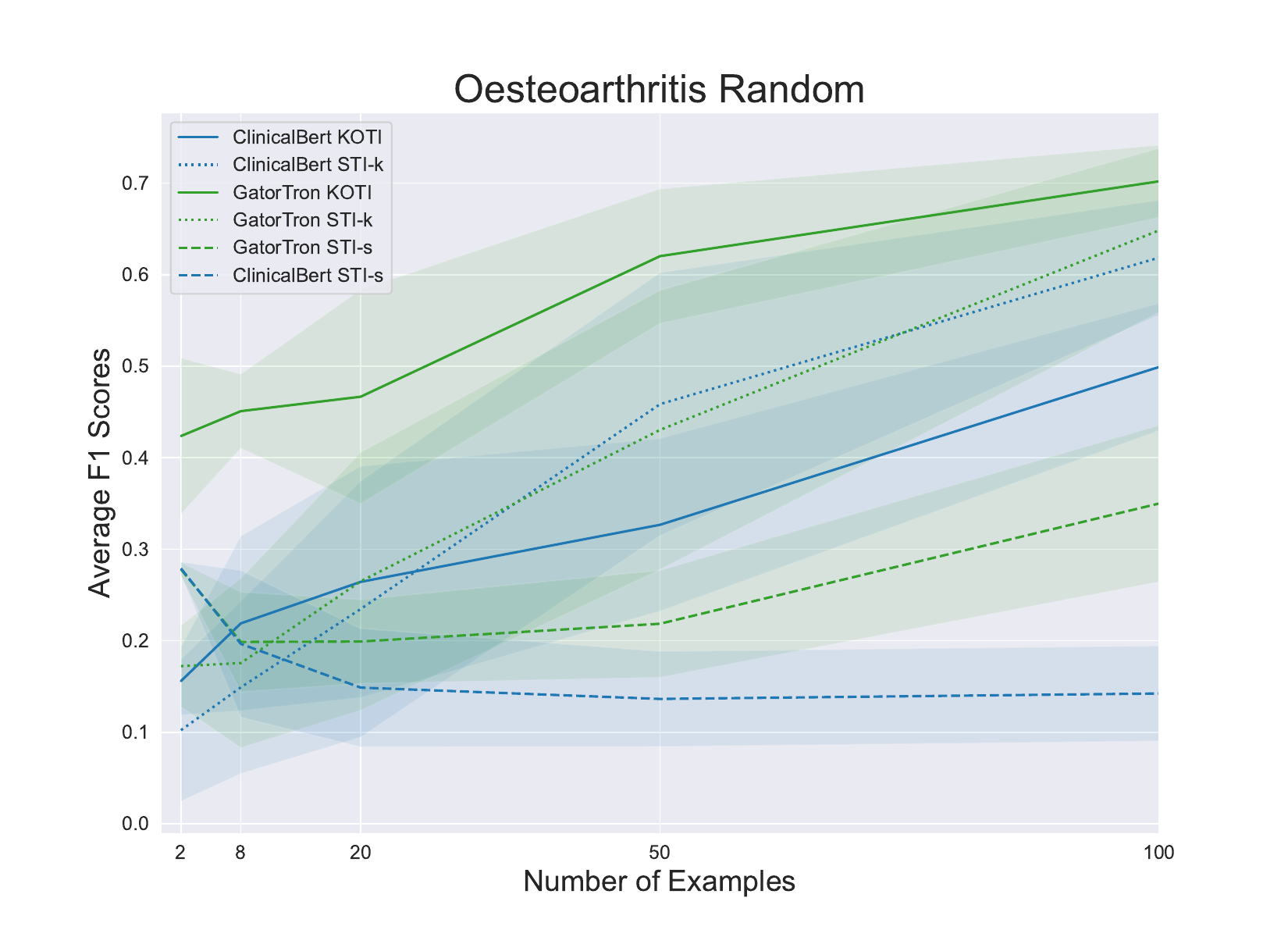}
\caption{Depression task with random examples}
\label{oa_r}
\end{figure}

We plot the average F1 scores as a function of number of training examples for each task and for balanced or random training examples separately in Figures \ref{dys_b} - \ref{oa_r}. 

Table \ref{token} Shows the average number of tokens per clinical notes in N2C2 obsity challenge, N2C2 smoking challenge and our dysmnorrhea dataset. It includes an estimate of the number of inference runs per clinical note required to process the whole input as described by  \citet{healthprompt} and used in the HealthPrompt framework. 

\begin{table*}[h]
\resizebox{1\textwidth}{!}{
\begin{tabular}{p{0.20\textwidth} p{0.20\textwidth} p{0.20\textwidth} p{0.20\textwidth} p{0.20\textwidth}}
\hline
\textbf{Data-set}               & \textbf{Average number of tokens per clinical note} & \textbf{SD}  & \textbf{Proportion of clinical notes with \textgreater 512 tokens} & \textbf{Estimated average number of inference runs per clinical note} \\
\hline
\textbf{N2C2 obesity challenge} & 1568                                       & 658 & 0.989                                                     & 3.1                                                          \\
\textbf{N2C2 smoking challenge }& 739                                        & 505 & 0.602                                                     & 1.4                                                          \\
\textbf{Dysmenorrhea}           & 1011                                       & 278 & 0.983                                                     & 2.0                          \\
\hline
\end{tabular}
}
\caption{Distribution of token length of clinical notes in different data-sets}
\label{token}
\end{table*}

\begin{figure}[h]
\includegraphics[width=0.5\textwidth]{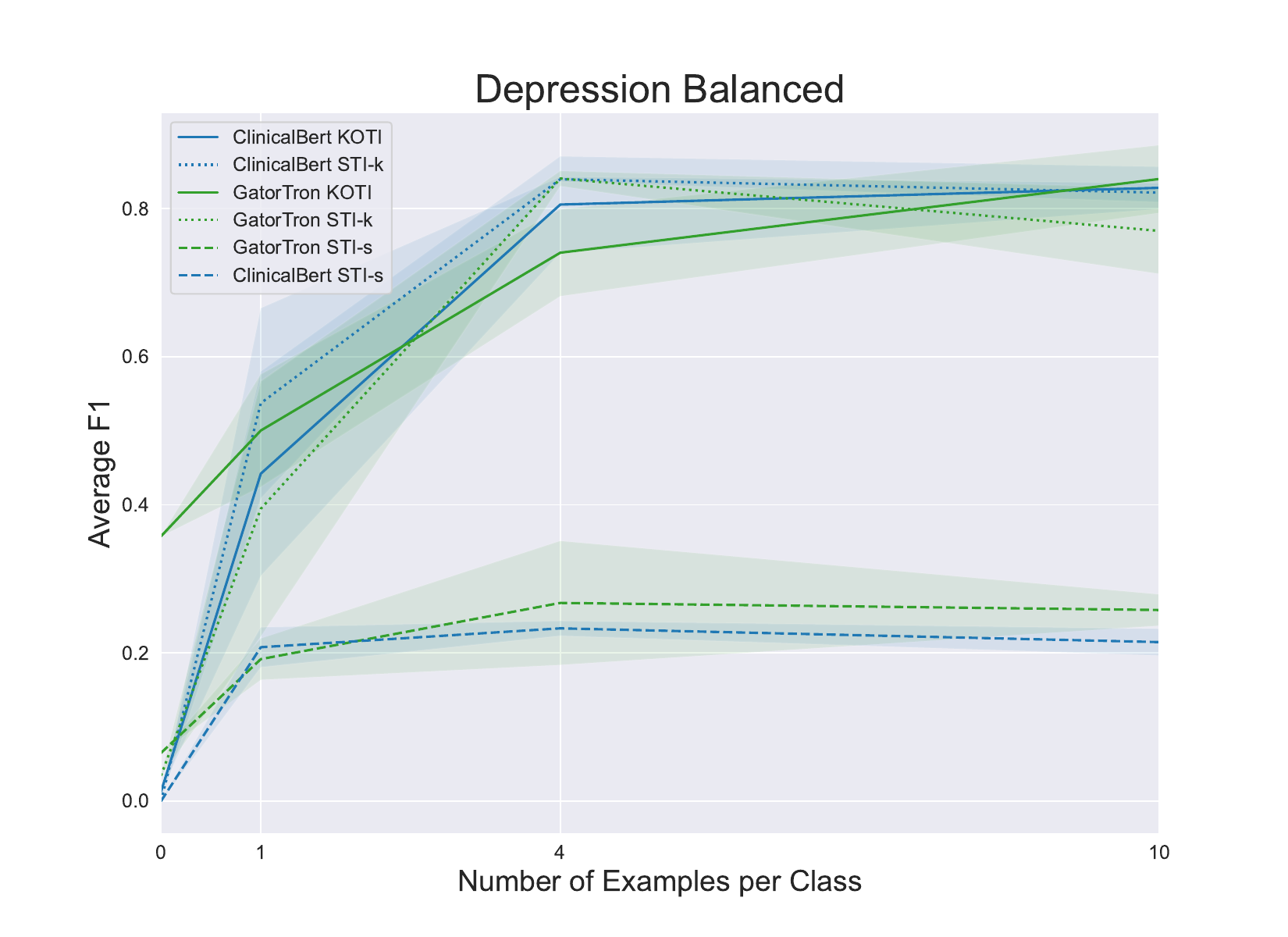}
\caption{Depression task with balanced examples}
\label{dep_b}
\end{figure}

\begin{figure}[h]
\includegraphics[width=0.5\textwidth]{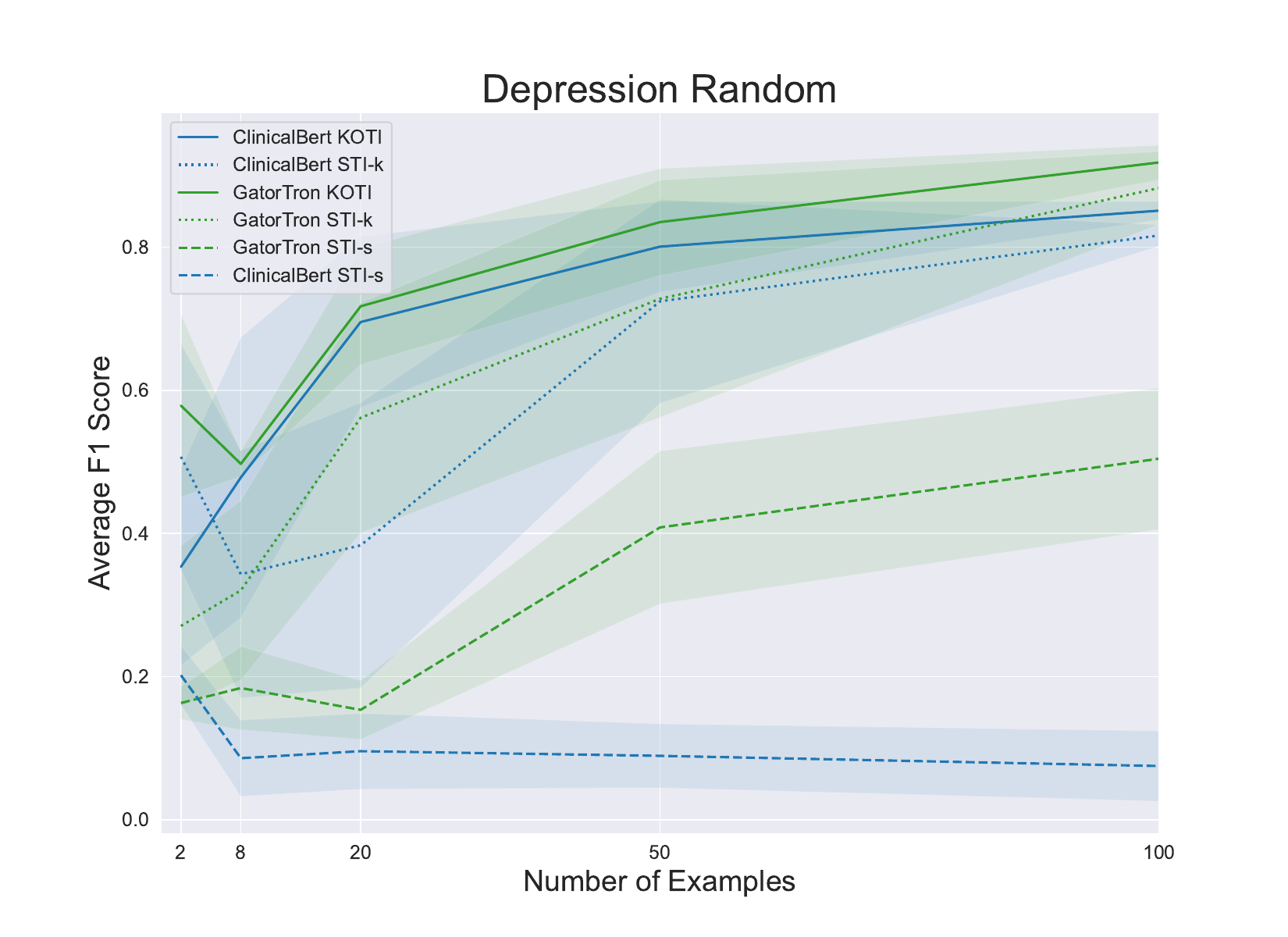}
\caption{Depression task with random examples}
\label{dep_r}
\end{figure}

\end{document}